\documentclass[letterpaper, 10 pt, journal, twoside]{ieeetran}  

\IEEEoverridecommandlockouts                              

\usepackage{tabularx}
\usepackage{hyperref}
\usepackage{xcolor}
\usepackage{cite}
\usepackage{booktabs}
\usepackage{graphicx}
\usepackage{caption}
\usepackage{array,booktabs,arydshln}

\definecolor{green}{RGB}{11,155,13}

\title{
Socially CompliAnt Navigation Dataset (SCAND): \\A Large-Scale Dataset Of Demonstrations For Social Navigation
}

\author{Haresh Karnan$^{1}$, Anirudh Nair$^{2}$, Xuesu Xiao$^{2}$, Garrett Warnell$^{2,5}$,\\ S\"oren Pirk$^{3}$, Alexander Toshev$^{3}$, Justin Hart$^{2}$, Joydeep Biswas$^{2}$ and Peter Stone$^{2, 4}$ 
\thanks{Manuscript received: February, 24, 2022; Revised May, 6, 2022;
Accepted May, 22, 2022.}
\thanks{This paper was recommended for publication by
Editor Dana Kulic upon evaluation of the Associate Editor and Reviewers’
comments. }
\thanks{$^{1}$The University of Texas at Austin, Department of Mechanical Engineering {\tt\small haresh.miriyala@utexas.edu}}%
\thanks{$^{2}$The University of Texas at Austin, Department of Computer Science, {\tt\small ani.nair@utexas.edu, \{xiao, joydeepb, hart, pstone\}@cs.utexas.edu}}%
\thanks{$^{3}$ Robotics@Google {\tt\small \{toshev, pirk\}@google.com}}%
\thanks{$^{4}$ Sony AI}%
\thanks{$^{5}$ Computational and Information Sciences Directorate, Army Research Laboratory {\tt\small garrett.a.warnell.civ@army.mil}}
\thanks{Digital Object Identifier (DOI): see top of this page.}
}

\begin{document}

\maketitle
\newcommand{\eg}{\textit{e.g.,}}

\begin{abstract}
Social navigation is the capability of an autonomous agent, such as a robot, to navigate in a ``socially compliant" manner in the presence of other intelligent agents such as humans. With the emergence of autonomously navigating mobile robots in human-populated environments (\eg{} domestic service robots in homes and restaurants and food delivery robots on public sidewalks), incorporating socially compliant navigation behaviors on these robots becomes critical to ensuring safe and comfortable human-robot coexistence. To address this challenge, imitation learning is a promising framework, since it is easier for humans to demonstrate the task of social navigation rather than to formulate reward functions that accurately capture the complex multi-objective setting of social navigation. The use of imitation learning and inverse reinforcement learning to social navigation for mobile robots, however, is currently hindered by a lack of large-scale datasets that capture socially compliant robot navigation demonstrations in the wild. To fill this gap, we introduce Socially CompliAnt Navigation Dataset (\textsc{scand})\textemdash a large-scale, first-person-view dataset of socially compliant navigation demonstrations. Our dataset contains 8.7 hours, 138 trajectories, 25 miles of socially compliant, human tele-operated driving demonstrations that comprises multi-modal data streams including 3D lidar, joystick commands, odometry, visual and inertial information, collected on two morphologically different mobile robots\textemdash a Boston Dynamics Spot and a Clearpath Jackal\textemdash by four different human demonstrators in both indoor and outdoor environments. We additionally perform preliminary analysis and validation through real-world robot experiments and show that navigation policies learned by imitation learning on \textsc{scand} generate socially compliant behaviors.

\end{abstract}

\section{INTRODUCTION}
\begin{figure}[!tb]
    \centering
    \includegraphics[scale=0.265]{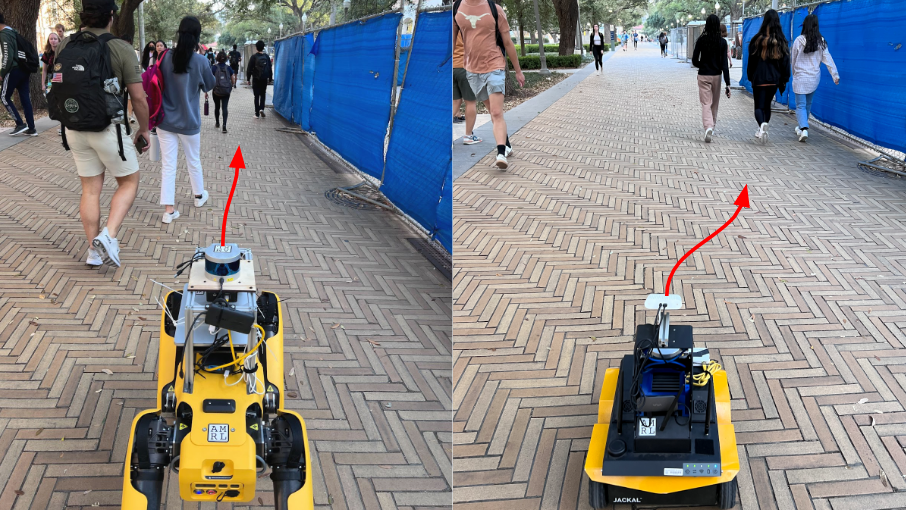}
    \caption{A human demonstrator teleoperates the two robots, following a socially compliant strategy (left- moving with traffic, right- sticking to the right of the road) around human crowds.}
    \label{fig:demo_speedway}
\end{figure}

\IEEEPARstart{S}{ocial}
 navigation is the capability of an autonomous agent to navigate in a socially compliant manner such that it recognizes and reacts to the objectives of other navigating agents, at least somewhat adjusting its own path in response, while also projecting signals that can help the other agents reciprocate. Enabling mobile robots to navigate in a socially compliant manner has been a subject of great interest recently in the robotics and learning communities \cite{socialforce, xuesusurvey, chen2018socially, collavoideverett, socialnavsurvey}. Towards enabling this capability, demonstration data of socially compliant navigation for mobile robots, such as the ones shown in Fig. \ref{fig:demo_speedway}, can be a valuable resource. For instance, such demonstration information can be used to learn socially compliant robot navigation using the paradigm of Learning from Demonstrations (\textsc{l}f\textsc{d}) \cite{lfdpaper, argall} or understanding human navigation in the presence of autonomous robots \cite{jackrabbot}.

Datasets for social navigation, generally used for learning and benchmarking, include data collected both in the real-world \cite{thorDataset2019} and in simulated environments \cite{tsoi2020sean, manso2020socnav1}. While such datasets provide basic trajectories of the robots and humans, they either contain limited interactions in constrained, orchestrated environments or restrict themselves to indoor-only navigation scenarios. When collecting data in such controlled settings \cite{thorDataset2019}, naturally occurring social interactions including---but not limited to---following lane rules of a country, yielding to pedestrians and vehicles, walking with and against a crowd of people, and street crossing is not captured.
Additionally, the robots used for data collection in previous social navigation datasets \cite{thorDataset2019} tend to use a simple controller for point-to-point navigation that does not explicitly exhibit socially aware navigation.

Recently, imitation learning has emerged as a useful paradigm for designing mobile robot navigation controllers \cite{byron_imitation_learning, voila, bojarski2016end, offroad_lecun}. In this paradigm, the desired navigation behavior is first demonstrated by an agent such as a human, the recording of which is then utilized by an imitation learning algorithm to imitate. This intuitive way of teaching a task to a robot is also easy for non-expert humans since it only requires providing demonstrations, instead of defining the rules of the task itself, which may be hard to explicitly define for social navigation. Motivated by recent successes of imitation learning in robot navigation, we posit that one way to enable autonomous agents to navigate socially is through learning from human demonstrations of socially compliant navigation behavior. However, there is a lack of large-scale datasets containing socially compliant navigation demonstrations in the wild that can be utilized for imitation learning.

To fill this gap, in this work, we introduce a dataset of demonstrations for socially compliant robot navigation in the wild. Our dataset contains 8.7 hours of human-teleoperated, socially compliant, navigation demonstrations, specifically, Velodyne lidar scans, joystick commands, odometry, camera visuals, and 6D inertial (IMU) information collected on two morphologically different mobile robots---a Clearpath Jackal and a Boston Dynamics Spot---within the University of Texas at Austin university campus. Comprising 25 miles in total of 138 trajectories, Socially CompliAnt Navigation Dataset (\textsc{scand}) is publicly released\footnote{\href{www.cs.utexas.edu/~xiao/SCAND/SCAND.html}{www.cs.utexas.edu/$\sim$xiao/SCAND/SCAND.html}} and also contains labeled tags of naturally occurring social interactions with every trajectory. 
Additionally, we demonstrate the utility of the dataset for studying questions relevant to social navigation. We first
show that there exists more than one strategy for an agent to navigate with social compliance, as it is possible for a classifier to differentiate between driving approaches of two different human demonstrators with an accuracy of $74.48\%$. Secondly, we also show that with \textsc{scand}, it is possible to learn socially compliant local and global navigation policies through imitation learning.




\section{RELATED WORK}
In this section, we review related literature with a focus on learning-based approaches for social navigation. We additionally survey relevant datasets for robot navigation and contrast their contributions with this work.

\subsection{Learning for Robot Navigation}

Recently, several algorithms have emerged that show the potential of applying learning to address challenges in robot navigation \cite{xuesusurvey}. Broadly speaking, in the robot navigation literature, learning-based approaches have been shown to be successful in problems such as adaptive planner parameter learning \cite{xiao2021appl}, overcoming viewpoint invariance in demonstrations \cite{voila}, and end-to-end learning for autonomous driving \cite{bojarski2016end, pfeiffer2018reinforced, wang2021agile}. Specifically in applying imitation learning for social navigation, the work by Tai et al. \cite{tai2018socially} is the closest to our work. They provide a simulation framework in gazebo along with a dataset generated using the same where virtual human agents navigate following the social force model \cite{socialforce}. They additionally train a social navigation policy using the Generative Adversarial Imitation Learning algorithm assuming the social force model as the ``expert" demonstrator and show a successful deployment of the learned policy in the real-world on a turtle bot robot. While their work has shown that imitation learning can be applied to address the social navigation problem, they do so assuming the social force model in simulation as the ``expert" demonstration. While simulated environments enable fast and safe data collection for online learning, they lack the naturally occurring social interactions seen in the wild. Also, as we show in section \ref{analysis}, there can be more than one strategy for an agent to navigate socially in a scene, which is not considered in their work. 

Other learning paradigms such as Reinforcement Learning (\textsc{rl}) have also been applied to address the social navigation problem. Everett et al. \cite{collavoideverett} present \textsc{ca}-\textsc{drl}, a multi-agent collision avoidance algorithm learned using \textsc{rl}. While this work shows impressive real-world results, their approach is limited to specific social scenarios and requires simulating these scenarios for the online learning algorithm to learn episodically. Kretzschmar et al. \cite{soc_com_robot_nav_paper} use Inverse Reinforcement Learning to learn cost functions for a socially compliant navigation policy. Similar to our work, they utilize human demonstrations of the social navigation task, however, they do so utilizing a small-scale, one-hour-long dataset. In this work, we contribute a large-scale dataset of robot social navigation demonstrations comprising multi-modal real-world data over multiple hours, both indoors and outdoors, on two different robots. Additionally, we train an imitation learning algorithm to show it is possible to learn socially compliant global and local navigation policies using our dataset.

\begin{table*}[]
\centering
\begin{tabular}{>{\centering\arraybackslash}m{1.2cm}>{\centering\arraybackslash}m{0.9cm}>{\centering\arraybackslash}m{1.3cm}>{\centering\arraybackslash}m{1.3cm}>{\centering\arraybackslash}m{5.2cm}>{\centering\arraybackslash}m{1.6cm}>{\centering\arraybackslash}m{1.6cm}>{\centering\arraybackslash}m{1.2cm}}
\toprule
\textbf{Dataset} & \textbf{\# Traj.} & \textbf{Dist. (Km)} & \textbf{Dur. (min)} & \textbf{Sensors}                                        & \textbf{Nav. method}  & \# \textbf{Robots} & \textbf{Location}             \\
\toprule
 CoBot \cite{biswas2013longterm} & 1082 & 131 & 15600 & 2D Range Scanner, RGB-D Camera, Wheel Odometry & Autonomous & 2 & Indoors + Outdoors \\
 \midrule
    L-CAS \cite{yz17iros} &
  3 &
  N/A &
  49 &
  3D LiDAR &
  Teleoperated &
  1 &
  Indoors \\
  \midrule
  NCLT \cite{carlevaris2016university} &
  27 &
  147.4 &
  2094 &
  3D LiDAR, RGB Camera, IMU, Wheel Odometry, GPS &
  Teleoperated &
  1 &
  Indoors + Outdoors \\
  \midrule
   
  FLOBOT \cite{zhimon2020jist} &
  6 &
  N/A &
  27.5 &
    3D LiDAR, RGB-D camera, Stereo Camera, 2D LiDAR, OEM incremental measuring wheel encoder, IMU &
    
    Autonomous &
  1 &
  Indoors \\
  \midrule

JRDB \cite{jackrabbot} &
  54&
  N/A&
  64&
  3D LiDAR, 2D LiDAR, Omnidirectional Stereo Suite, RGB camera, RGB-D stereo camera, 6D IMU &
    Teleoperated &
  1 &
  Indoors + Outdoors \\
  \midrule

TH{\"O}R \cite{thorDataset2019} &
  600 &
  N/A &
  60 &
  3D LiDAR, Motion capture system, Eye-tracking Glasses &
  Autonomous &
  1 &
  Indoors \\
  \midrule
  
  SCAND &
  138 &
  40 &
  522 &
  3D LiDAR, RGB-D Camera, Monocular Camera, Stereo Camera, Wheel Odometry, Visual Odometry &
  Teleoperated &
  2 &
  Indoors + Outdoors \\
\bottomrule

\end{tabular}
\caption{Comparison of real-world datasets for robot navigation.}
\label{existing_datasets}
\end{table*}

\subsection{Datasets for Social Navigation}
Over the last decade, datasets containing robots navigating in both simulated and real-world environments have been useful for a wide variety of research areas, such as tracking groups of people \cite{thorDataset2019, lau2009track, linder2016multi}, human trajectory prediction \cite{chung2012incremental}, navigation \cite{hirose2019deep}, robot localization \cite{biswas2013longterm, kitti, biwasenml} and collision risk assessment \cite{lo2019robust}.

\subsubsection{Simulated Datasets for Social Navigation}
Social environments in simulation can provide researchers with fast data collection on social navigation \cite{tsoi2020sean, tsoi2021approach, holtz2021socialgym, tai2018socially}. Moreover, such simulated environments can be generated with a specified number of elements: the number and locations of the humans, the structure of the room, the number of objects, and the interactions between people and between objects and people \cite{manso2020socnav1}. While simulated platforms provide these benefits, they are limited in that they lack the natural, real-world interactions that are experienced by humans. Datasets that capture real-world robot navigation data in the wild provide researchers with more naturally occurring scenarios \cite{biswas2013longterm,zhimon2020jist,yz17iros,carlevaris2016university}. Additionally, datasets collected in the wild provide sensory data for these scenarios which can be then used for perceptual tasks related to navigation \cite{de2021deepsocnav}.

\subsubsection{Real-world Datasets for Robot Navigation}

In addition to simulated datasets, several real-world datasets for long-term robot navigation in human environments have also been made available over the last decade. In the CoBots dataset \cite{biswas2013longterm}, two CoBots we deployed indoors autonomously using a topological graph planner and collected more than 130 km worth of laser scans, odometry, and localization data over 1082 deployments. Similarly, the L-CAS \cite{yz17iros}, FLOBOT \cite{zhimon2020jist}, JRDB \cite{jackrabbot} and NCLT \cite{carlevaris2016university} datasets contain LiDAR scans, RGBD visuals, GPS, and IMU data collected independently on different robots, addressing perception-related challenges to long-term robot navigation. In all these different datasets, the robots were deployed in a public environment, such as a restaurant or a university campus, and teleoperated by a human as opposed to being autonomous, but these teleoperated demonstrations are not explicitly socially compliant. The JRDB social navigation dataset \cite{jackrabbot} is the closest to our work, but it is smaller in scale, containing only 64 minutes worth of data from 54 indoor and outdoor trajectories. While the focus of the JRDB dataset is to solve perception-related challenges such as human tracking and detection in social navigation, the focus of the \textsc{scand} dataset in this work is to address the ``navigation" sub-component of social navigation. The TH{\"O}R dataset \cite{thorDataset2019} provides motion trajectories of both robots and humans using tracking helmets. However, this is smaller in scale since it contains only one hour's worth of data. Also, the data is collected indoors in an 8.4x18.8m laboratory room with an orchestrated social navigation scenario for the human agents in the scene and a socially unaware, pre-defined path for the robot---adjusting neither its speed nor trajectory to account for surrounding people. Existing real-world datasets for robot navigation are summarized in Table \ref{existing_datasets}.

While previous datasets collected with robots and humans have proven to be useful to study localization, perception, and other navigation-related challenges, they lack demonstration information in the form of motion commands and navigation strategies in different social scenarios that could help us understand socially compliant robot navigation in the presence of other autonomous agents. The \textsc{scand} dataset introduced in this work addresses this gap and provides rich human demonstration information in the form of joystick commands and multi-modal robot sensor data in different, naturally occurring social scenarios. \textsc{scand} also contains labeled tags of twelve different social interactions that occurred along the path. Also, since robots of different morphologies and capabilities could navigate differently and induce different social interactions, \textsc{scand} also includes data from two different robots. For example, the legged Spot, capable of climbing stairs could choose to prefer the stairs along its path while navigating whereas the wheeled Jackal might choose a ramp to navigate. The other datasets use only one robot to collect data (the Cobots dataset \cite{biswas2013longterm} uses two robots but they are morphologically the same). Using two morphologically different robots makes \textsc{scand} useful to investigate social navigation in robots with different morphologies (wheeled vs. legged).

\begin{figure*}[!tb]
    \centering
    \includegraphics[scale=0.445]{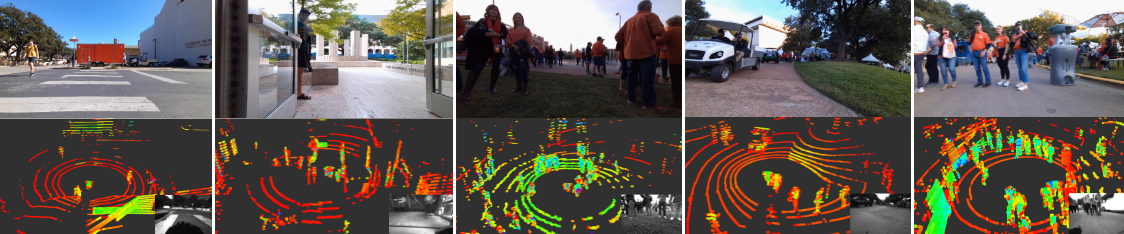}
    \caption{Five example scenarios from \textsc{scand} showing the RGB image and below it the accompanying Lidar with the monocular image from side camera on the Spot. From left to right, the scenarios have the tags ``Street Crossing", ``Narrow Doorway, ``Navigating Through Large Crowds", ``Vehicle Interaction", and ``Crossing Stationary Queue."}
    \label{fig:intersting_social_interactions}
\end{figure*}

\section{DATA COLLECTION PROCEDURE}
In this section, we first describe the data collection procedure used in \textsc{scand} and outline the sensor-suite present on both robots. We then describe the labeled annotations of social interactions provided with every trajectory.

\subsection{Collecting Data}
To collect multi-modal, socially compliant demonstration data for robot navigation, four human demonstrators---including the first two authors of this work---navigate the robot by teleoperation using a joystick. We collected data within the UT Austin university campus, with the demographics of the humans in the scene comprised mostly of students, faculty, and other campus denizens.
For each of the 138 trajectories in \textsc{scand}, the human demonstrator walks behind the robot at all times, maintaining on average two meters distance. The human demonstrator does not explicitly interact with the crowd in the scene. Unlike other datasets for social navigation \cite{thorDataset2019}, we do not restrict data collection to a controlled, indoor environment or orchestrate a social scenario for data collection. Instead, similar to the JRDB dataset \cite{jackrabbot}, we perform data collection in the wild in both indoor and outdoor environments.  The two robots are driven around the university campus on frequently used sidewalks, roads, and lawns, and inside buildings, all with people in the scene during peak hours of high foot traffic. This includes data collected outdoors near the university's football stadium on two game days with high traffic public crowds gathered near the arena. The Spot is driven at linear and angular velocities in the range of $[0, 1.6]$ $m/s$ and $[-1.5, 1.5]$ $rad/s$, respectively, and the Jackal in the range of $[0, 2.0]$ $m/s$ and $[-1.5, 1.5]$ $rad/s$, respectively. Note that these velocities are within the range of many people's normal walking speed.  

\begin{figure}[!tb]
    \centering
    \includegraphics[scale=0.22]{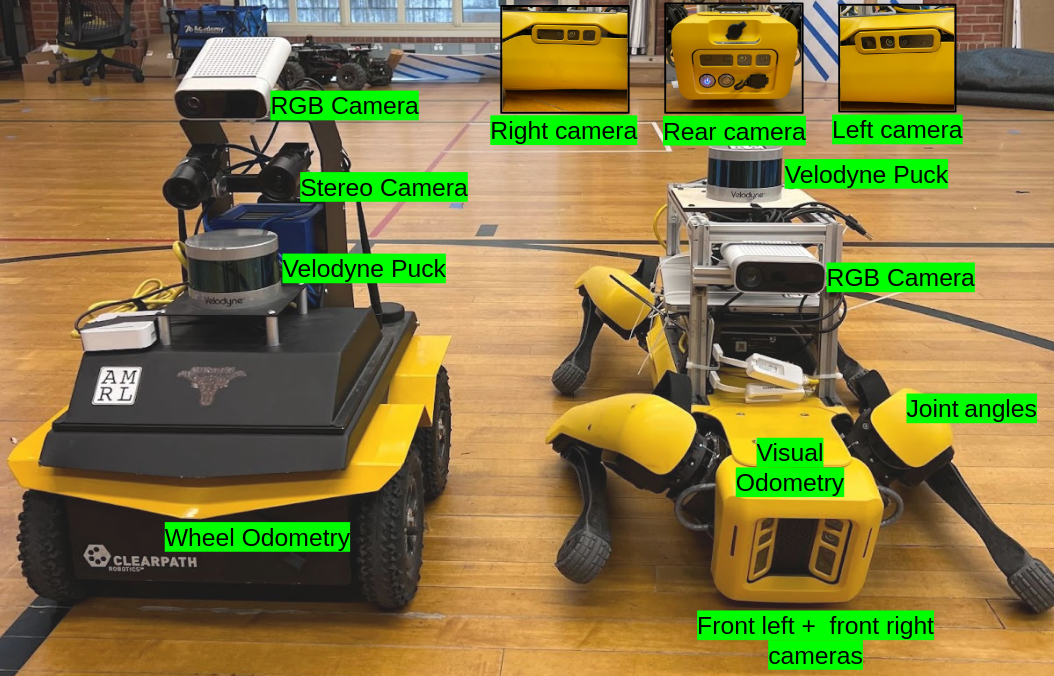}
    \caption{Sensors present on the wheeled Jackal and the legged Spot robots. Along with this multi-modal sensor information, \textsc{scand} also contains joystick commands issued during the navigation demonstration.}
    \label{fig:sensor_suite}
\end{figure}

Fig. \ref{fig:sensor_suite} shows the sensors present on the Clearpath Jackal and the Boston Dynamics Spot robots. Both robots have in common a VLP-16 Velodyne laser puck publishing at a frequency of 10 Hz, a 6D inertial (IMU) sensor at 16 Hz, and a front-facing Azure Kinect RGB camera at 20 Hz. In addition to these common sensors, the Jackal has a front-facing stereo camera (20 Hz) and wheel odometry (30 Hz), while the Spot has five monocular cameras on its body (publishing at 5 Hz), placed as shown in Fig. \ref{fig:sensor_suite}. We utilize the Boston Dynamics APK to record the visual odometry of its body frame and the joint angles of the legs on the robot. \textsc{scand} also contains transforms between the frames of each of the sensors relative to the robot's body for both robots. We utilize AMRL's software stack \cite{graphnavgithub} for data collection from different sensors which we record in the rosbag format \cite{ros}.

Although we provide visual information of the scene in the form of surround-view monocular images on the Spot, RGB image from the front-facing Kinect camera, and 3D Velodyne laser scans on both robots, since the focus of this work is specifically on navigation, we do not provide any labeled annotations for human detection or tracking. We refer the reader to the JRDB dataset \cite{jackrabbot} which contains detailed, high-quality annotations for solving perception-related tasks. Instead, \textsc{scand} contains joystick commands of linear and angular velocities executed by the demonstrator while teleoperating the robot socially, along with rich, multi-modal sensory information of the environment including labeled annotations of 12 different social interactions in every trajectory. Fig. \ref{fig:intersting_social_interactions} shows five example scenarios and their associated tags.

\begin{table}[!h]
\centering
\begin{tabular}{>{\centering\arraybackslash}m{0.1\textwidth}>{\centering\arraybackslash}m{0.25\textwidth}>{\centering\arraybackslash}m{0.05\textwidth}}
\toprule
 \textbf{Tag} & \textbf{Description} &
 \textbf{\# Tags} \\
 \toprule
 Against Traffic & Navigating against oncoming traffic & 22 \\ 
 \midrule
 With Traffic & Navigating with oncoming traffic & 74 \\
 \midrule
 Street Crossing & Crossing across a street & 34 \\
 \midrule
 Overtaking & Overtaking a person or groups of people & 14 \\
 \midrule
  Sidewalk & Navigating on a sidewalk & 57 \\
  \midrule
  Passing Conversational Groups & Navigating past a group of 2 or more people that are talking amongst themselves & 38 \\
  \midrule
  Blind Corner & Navigating past a corner where the robot cannot see the other side & 6 \\
  \midrule
  Narrow Doorway & Navigating through a doorway where the robot waits for a human to open the door & 15 \\
  \midrule
  Crossing Stationary Queue & Walking across a line of people & 6 \\
  \midrule
  Stairs & Walking up and/or down stairs & 22 \\
  \midrule
  Vehicle Interaction & Navigating around a vehicle & 21 \\
  \midrule
 Navigating Through Large Crowds & Navigating among large unstructured crowds & 27 \\ 
\bottomrule
\end{tabular}
\caption{Descriptions of labeled tags contained in \textsc{scand}}

\label{table1} 
\end{table}

\subsection{Labeled Annotations of Social Interactions}
We annotate each trajectory in \textsc{scand} with labels describing social interactions that occurred along the path. The labels are in the form of a list of textual captions of social interactions taking place in a trajectory, chosen from a set of twelve predefined labels of social interactions observed in \textsc{scand}. For the full list of labels, refer to Table ~\ref{table1}. We intend the labels to be useful for future studies of specific scenarios that occur during social navigation in the real-world. 

\section{ANALYSIS}
\label{analysis}

In this section, we provide analysis on \textsc{scand} with the data collected on the Spot to illustrate the usefulness of this dataset for answering a variety of questions related to social navigation. Specifically, we ask the following questions: 

\begin{enumerate}
    \item Is there more than one strategy for socially navigating in a scene? 
    \item Can we learn a local and global planner for social navigation using \textsc{scand} ?
\end{enumerate}

We answer question 1 in subsection \ref{demonstrator_classification} by learning a neural network-based classifier that is trained for the task of demonstrator classification given a ten-second sequence of sensor observations and joystick commands as input. We then answer question 2 in subsection \ref{imitation_learning} by applying the behavior cloning (\textsc{bc}) imitation learning algorithm \cite{behaviorcloning} on \textsc{scand} to learn a global and local planner jointly.


\begin{figure}[!tb]
    \centering
    \includegraphics[scale=0.30]{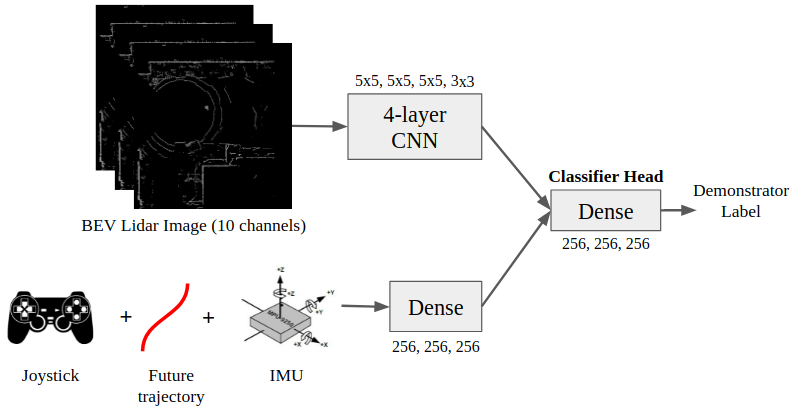}
    \caption{Network architecture and inputs for the demonstrator classifier. The classifier takes as its input ten-second long sensor observations and predicts a demonstrator label. The \textsc{bc} agent (not shown in this figure) follows a similar architecture, with a global planner and local planner head instead of a classifier head. Additionally, instead of the future trajectory, the \textsc{bc} agent takes as its inputs the \texttt{move\textunderscore base} global plan and desired velocities.}
    \label{fig:classifier_arch}
\end{figure} 

\subsection{Demonstrator Classification}
\label{demonstrator_classification}

In this subsection, we consider the question ``is there more than one strategy for socially navigating in an environment?" We hypothesize that the answer is yes, there is more than one strategy to navigate in a socially compliant manner in a given scenario.

\subsubsection{Approach and Implementation}

To answer this question and to validate our hypothesis, we choose sixteen trajectories driven by two demonstrators navigating along the same route (Speedway road within the university campus) and train a neural network for the task of demonstrator classification (training on twelve trajectories and validating on four trajectories). The input to our classifier is a ten-second long sequence of sensor observations. This sequence consists of processed sensor observations provided in \textsc{scand} such as lidar scans (subsampled to 1 Hz and represented as grayscale bird's eye view (BEV) image), positions of the robot relative to the first lidar frame, future trajectory driven by the human consisting of 200 points in the most recent lidar frame, inertial and joystick values executed by the demonstrator at each of the lidar frames. The neural network architecture consists of a four-layer convolutional encoder to process the grayscale BEV lidar images and a three-layer fully connected network to process the other sensor observations. The representations output by these layers are fed into a three-layer fully connected network classifier head. We use the binary cross-entropy loss to train the classifier network. Fig. \ref{fig:classifier_arch} shows the inputs and neural network architecture of the demonstrator classifier. 
\subsubsection{Results and Conclusion}
We find that the classifier is 74.48\% accurate at classifying the expert on the held-out test set. Given that random guessing would lead to a success rate of 50\%, and that many ten-second trajectories do not indicate any differentiating social interactions, this number is indicative of successful prediction. 
The ability of the classifier to identify the demonstrator from their navigation style with an accuracy of 74.48\% using a ten-second sequence of observations, combined with the fact that the demonstrations in \textsc{scand} are socially compliant shows that there exists more than one strategy for socially compliant navigation in a given scenario, validating our hypothesis. Enabling algorithms to take into consideration this manifold of socially compliant robot navigation behaviors naturally observed in humans demonstrations is a promising direction for future work.

\subsection{Imitation Learning for Global and Local Planning}
\label{imitation_learning}

\subsubsection{Approach and Implementation}

To answer question 2, we apply the \textsc{bc} imitation learning algorithm \cite{behaviorcloning} on \textsc{scand} to jointly train end-to-end a socially-aware global and local planner for robot navigation. The objective of the global planner agent is to predict the socially compliant global plan (the future trajectory driven by the human demonstrator, within a horizon of ten meters distance from the robot). The local planner agent's objective is to predict the forward and the angular velocities demonstrated in \textsc{scand} in a socially compliant manner. We jointly train the local and the global planner agents using a common representation space of observations, similar to the demonstrator classifier network shown in Fig. \ref{fig:classifier_arch}. However, unlike the demonstrator classifier network with a single classifier head, here we use two different heads (three-layer fully connected networks) for the global and the local planner agents.
As inputs to the \textsc{bc} agent, we provide processed sensor observations from \textsc{scand} of two seconds in length to account for temporal variations in the scene; this includes BEV lidar scans (subsampled to 2 Hz and represented as grayscale BEV image as shown in Fig. \ref{fig:classifier_arch}), positions of the previous lidar frames relative to the first lidar frame and inertial information at each of the lidar frames. Additionally, we also provide the global path and desired velocities produced by \texttt{move\textunderscore base} \cite{rosmovebase} using the location of the robot ten meters in the future from its current position as prior information to the network. We posit that feeding this prior information from \texttt{move\textunderscore base} as inputs to the \textsc{bc} agent would enable improved performance. The global planner head predicts 200 points in the path driven by the demonstrator, and the local planner predicts 20 timesteps of joystick commands ($v$, $\omega$) issued by the demonstrator since the current frame. We sum the mean-squared error loss objectives for both agents and update their parameters together. Note that we do not utilize any representation learning algorithm to pretrain the encoders that process the sensor observations, but doing so may potentially improve results. However, since the focus of this analysis is to show the potential of \textsc{scand} in enabling existing imitation learning algorithms to learn socially compliant navigation policies, representation learning is left to future work.

\begin{figure}[!tb]
    \centering
    \includegraphics[scale=0.52]{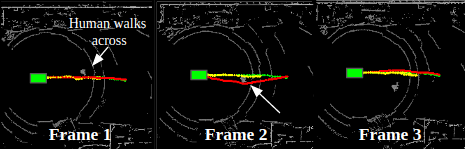}
    \caption{An example sequence of three BEV lidar frames of a human walking across the robot's (green box) path. Green path shows the demonstrated trajectory, red path shows the \texttt{move\textunderscore base} global path, and the yellow path shows the predicted trajectory by the learned \textsc{bc} agent. In frame 2, the \texttt{move\textunderscore base} path moves in the direction of the human's future state, whereas the learned path closely follows the desired socially compliant path.}
    \label{fig:paths}
\end{figure} 

\subsubsection{Results and Conclusion}

To evaluate the social navigation behavior of the global planner, we compute the Hausdorff distance metric on a held out test set, between the global path predicted by the learned global planner agent and the actual path driven by the demonstrator in the future. The average Hausdorff distance between the \texttt{move\textunderscore base} global path and the demonstrated path in a held out test set is 1.25. However, after training the \textsc{bc} global planner agent on \textsc{scand}, the average Hausdorff distance between the predicted trajectory and the demonstrated trajectory is improved at 0.26. Fig. \ref{fig:paths} shows a scenario involving the robot, and a human walking across the robot's path. We see that in this scenario, the predicted path closely matches that of the socially compliant demonstrated path, whereas \texttt{move\textunderscore base} turns in the direction of the human's future state, creating an undesired interaction. 

\begin{figure}[!tb]
    \centering
    \includegraphics[scale=0.24]{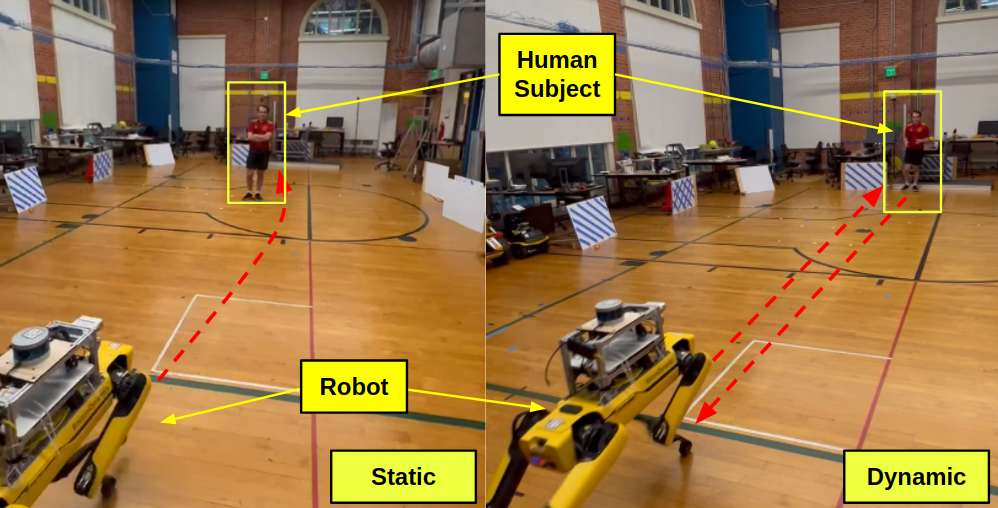}
    \caption{Evaluating the local planner agent trained using Behavior Cloning on \textsc{scand}. Scenario on the left shows a stationary human in the robot's path and the scenario on the right shows a human walking to the location of the robot. The robot is evaluated on social compliance and safety as it navigates to its goal position.}
    \label{fig:human_trial_setup}
\end{figure} 

\begin{figure}[!tb]
    \centering
    \includegraphics[scale=0.35]{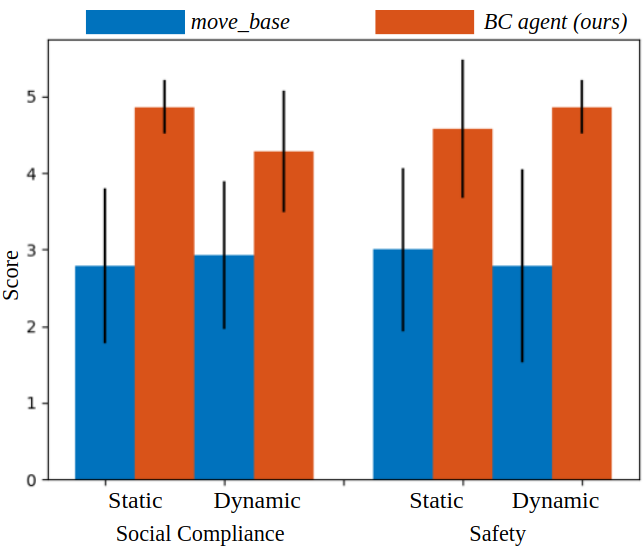}
    \caption{Mean and standard deviation of scores assigned by the fourteen human participants in the evaluation study for the learned local planner. }
    \label{fig:human_trial_metric}
\end{figure}

To validate the learned local planner agent, we conduct real-world experiments using the Spot robot with fourteen human participants in an indoor location. We design two scenarios---static and dynamic---to evaluate the social compliance and safety of the learned local planner and the \texttt{move\textunderscore base} planner, as shown in Fig. \ref{fig:human_trial_setup}. In the static scenario, the robot starts five meters ahead of a stationary human in the robot's path, and tries to navigate to a goal position five meters behind the human. In the dynamic scenario, the robot and the human start facing each other 10 meters apart and try to reach the start position of the other. In the dynamic scenario, the participants were asked to navigate in a socially compliant manner to their goal position and in both scenarios, the participants were asked to observe the navigation behavior of the robot. After each scenario, for both the algorithms, a questionnaire was presented to the participant with the two following questions:
\begin{enumerate}
    \item \textit{On a scale of 1 to 5, how ``socially compliant" do you think the robot was? (think of social compliance as how considerate the robot was of your presence)}
    \item \textit{On a scale of 1 to 5, how ``safe" did you feel around the robot?}
\end{enumerate}
We randomized the order in which the two algorithms (\texttt{move\textunderscore base} and \textsc{bc} policy) were played to the participants. Fig. \ref{fig:human_trial_metric} shows the responses of the human participants. On average, more humans felt the imitation learning agent trained on \textsc{scand} was more socially compliant (\textsc{scand} mean=$4.39$, sd=$0.99$; \texttt{move\textunderscore base} mean=$2.86$, sd=$0.82$) and safer (\textsc{scand} mean=$4.71$, sd=$0.70$; \texttt{move\textunderscore base} mean=$2.89$, sd=$1.18$) than the \texttt{move\textunderscore base} agent. The results for both questions are statistically significant as tested by a One-Way Analysis of Variance (ANOVA) (\textit{Safe} $F_{1,55}=47.87, p < 0.001$; \textit{Socially Compliant} $F_{1,55}=38.67, p < 0.001$). This is expected since the \texttt{move\textunderscore base} agent is not designed to exhibit social compliance. Refer to the attached supplementary video for scenarios showing the behavior of both the algorithms in the static and dynamic trials. The results of this study support our hypothesis that imitation learning using demonstrations provided in \textsc{scand} produces socially compliant navigation policies. In the interest of reproducibility, we provide the 75\%-25\% train-test splits of the trajectories collected using the Spot robot in \textsc{scand}. 

While we show here that the \textsc{bc} agent is able to handle simple social navigation scenarios, better imitation learning algorithms may be needed to handle more sophisticated social navigation scenarios that are present in \textsc{scand}. 

\section{ANTICIPATED USE CASES}
 Although \textsc{scand} includes a wide variety of social navigation scenarios, there may be novel interactions that are less frequent. To improve generalizability of a learning based approach to unseen situations, exploring representation learning for social navigation with \textsc{scand} is a promising future direction. \textsc{scand} was collected in a single city (Austin, Texas, USA) and might incorporate regional biases such as staying to the right of the road, or overtaking pedestrians from the left. This potential bias raises a need for algorithms, evaluations, and metrics for social navigation that are flexible enough to work in the presence of different local norms.

Evaluating social navigation policies is an active area of research in the navigation community \cite{pirk2022protocol, tsoi2020sean, socmomentum, biswas2021socnavbench}. While benchmarking social navigation policies is out of scope for this paper, existing simulation-based navigation benchmarks such as SocNavBench \cite{biswas2021socnavbench} that use human-only trajectories could be augmented and improved using human-robot interaction trajectories in \textsc{scand}. Similarly, another interesting future research direction is to explore Real-to-Sim transfer \cite{sim2real, garat, sgat, rgat} with \textsc{scand} and improve parameterized simulated social navigation environments to generate more realistic social interactions between virtual agents, directly benefiting data hungry approaches such as reinforcement learning.


Other directions for future work that could directly benefit from \textsc{scand} include trajectory prediction, trajectory classification, and inverse reinforcement learning for large-scale cost function learning. Previously, work on trajectory prediction and classification has used human-only \cite{humantrajpred} or robot-only \cite{robotmotionpred} trajectories, but with access to \textsc{scand}, exploring human-robot trajectories is an interesting direction for future work. The work by Wulfmeir et al. \cite{medirl} utilized static scenarios to learn a cost function for autonomous robot navigation using Maximum Entropy Deep Inverse Reinforcement Learning (\textsc{medirl}). Applying \textsc{medirl} on \textsc{scand} to learn cost functions that incorporate social compliance is also an interesting direction for future work.  
 
 \section{CONCLUSION}

In this work, we introduce the Socially CompliAnt Navigation Dataset (\textsc{scand}), a large-scale dataset of demonstrations for mobile robot social navigation. \textsc{scand} contains 8.7 hours, 138 trajectories, 25 miles of socially compliant driving demonstrations, collected on two morphologically different robots. In addition to the multi-modal sensory data streams from the two robots, \textsc{scand} also contains labeled annotations of social interactions for all trajectories. We illustrate the usefulness of \textsc{scand} for answering a variety of questions related to social navigation. First, we show that there exists more than one strategy for socially compliant navigation by training a neural network classifier on the task of demonstrator classification. Second, we train a behavior cloning agent on the demonstrations from \textsc{scand} and show that it is possible to learn both a socially compliant global and local planner for mobile robot navigation using \textsc{scand}. We further validate the performance of the behavior cloned local planner through human trials on two social navigation scenarios and show that the participants perceived the imitation learning agent to be relatively more socially compliant and safe, compared to a naive \texttt{move\textunderscore base} agent. 
 
\section*{ACKNOWLEDGMENTS}
\small
This work has taken place in the Learning Agents Research
Group (LARG) and Autonomous Mobile Robotics Laboratory (AMRL) at UT Austin. LARG research is supported
in part by NSF (CPS-1739964, IIS-1724157, NRI-1925082),
ONR (N00014-18-2243), FLI (RFP2-000), ARO (W911NF19-2-0333), DARPA, Lockheed Martin, GM, and Bosch.
AMRL research is supported in part by NSF (CAREER2046955, IIS-1954778, SHF-2006404), ARO (W911NF-19-2-
0333, W911NF-21-20217), DARPA (HR001120C0031), Amazon, JP Morgan, and
Northrop Grumman Mission Systems. Peter Stone serves as
the Executive Director of Sony AI America and receives financial compensation for this work. The terms of this arrangement
have been reviewed and approved by the University of Texas at
Austin in accordance with its policy on objectivity in research. Human subjects research covered under The University of Texas at Austin IRB Number STUDY00002561 and STUDY00002562. 

\bibliographystyle{IEEEtran}
\bibliography{mybib}

\end{document}